\pdfoutput=1

\documentclass[11pt]{article}

\usepackage{EMNLP2022}

\usepackage{times}
\usepackage{latexsym}

\usepackage[T1]{fontenc}

\usepackage[utf8]{inputenc}

\usepackage{microtype}

\usepackage[namelimits]{amsmath} 
\usepackage{amssymb}            
\usepackage{amsfonts}           
\usepackage{mathrsfs}   
\usepackage{multirow}           
\usepackage{comment}
\usepackage{graphicx} 
\usepackage{bm}
\usepackage{booktabs}
\usepackage{makecell}
  
\usepackage{tabularx}

\usepackage{algorithm}
\usepackage{algpseudocode}

\usepackage{inconsolata}

\title{Long Text and Multi-Table Summarization: Dataset and Method}

\author{Shuaiqi Liu, Jiannong Cao, Ruosong Yang, and Zhiyuan Wen\\
  The Hong Kong Polytechnic University\\
  \texttt{\{cssqliu, csjcao, csryang, cszwen\}@comp.polyu.edu.hk} \\}

\begin{document}
\maketitle

\begin{abstract}
Automatic document summarization aims to produce a concise summary covering the input document's salient information. Within a report document, the salient information can be scattered in the textual and non-textual content. However, existing document summarization datasets and methods usually focus on the text and filter out the non-textual content. Missing tabular data can limit produced summaries' informativeness, especially when summaries require covering quantitative descriptions of critical metrics in tables. Existing datasets and methods cannot meet the requirements of summarizing long text and multiple tables in each report. To deal with the scarcity of available data, we propose FINDSum, the first large-scale dataset for long text and multi-table summarization. Built on 21,125 annual reports from 3,794 companies, it has two subsets for summarizing each company's results of operations and liquidity. To summarize the long text and dozens of tables in each report, we present three types of summarization methods. Besides, we propose a set of evaluation metrics to assess the usage of numerical information in produced summaries. Dataset analyses and experimental results indicate the importance of jointly considering input textual and tabular data when summarizing report documents.
\end{abstract}

\section{Introduction}

\begin{figure}[t]
\centering
\includegraphics[width=2.8in]{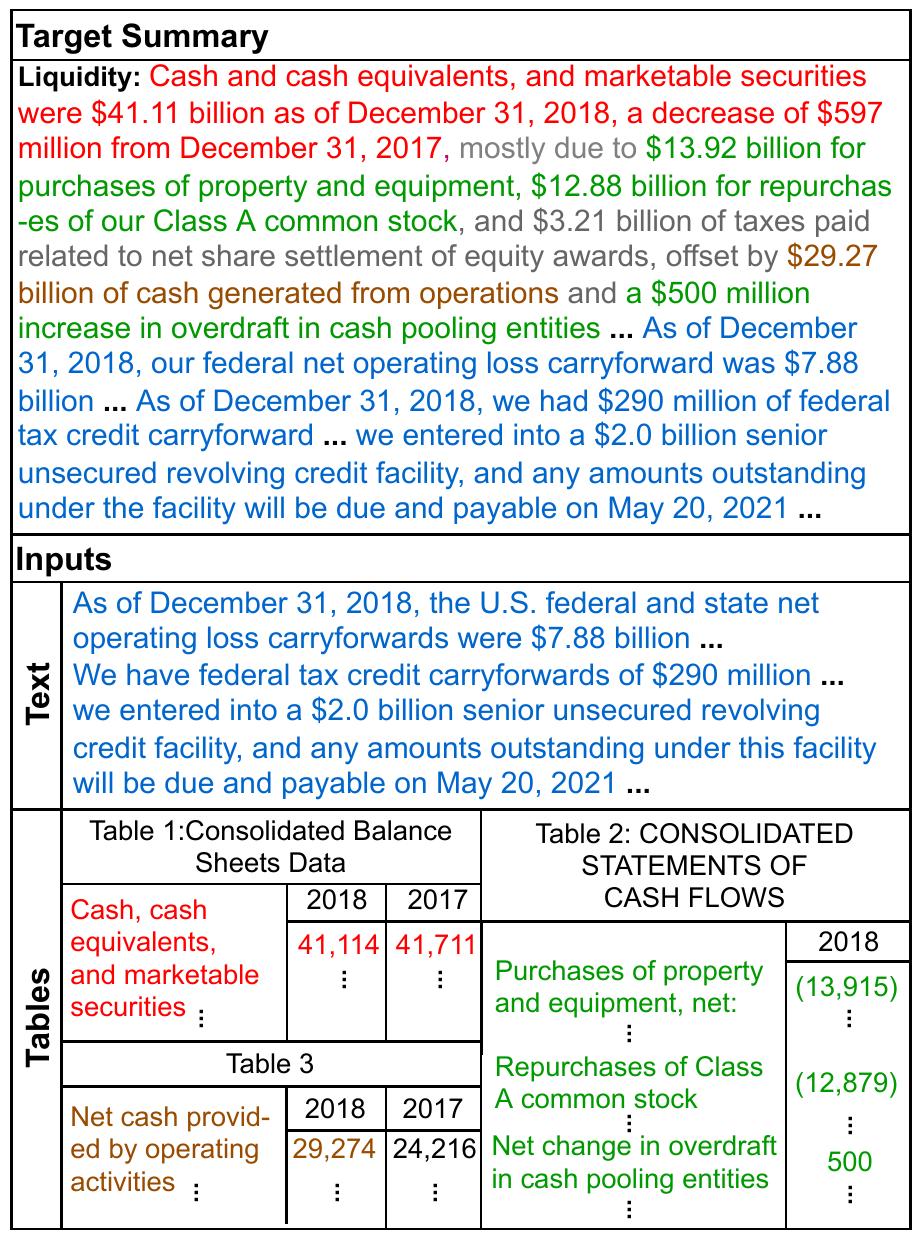}
\caption{An example from the FINDSum dataset. The content found in the target summary is color-coded.}\label{fig:FindSum_example} 
\end{figure} 

Report documents, like financial reports, investigative reports, and technical reports, are essential information sources.
These reports usually contain textual and tabular content.
As shown in Figure \ref{fig:FindSum_example}, the salient information can be scattered in long text and multiple tables in each report, which makes it difficult for non-specialized readers to efficiently read and gather salient information from these report documents.  
Automatic document summarization techniques can produce these reports' summaries, which can support readers quickly browsing salient information in these reports. 
Our target is to let the computer generate an informative, fluent, and non-redundant summary for the long text and multiple tables in each report. 
To achieve this target, we need to deal with some challenging issues: the scarcity of available data, identifying the salient information scattered in a large amount of input content, incorporating different types of content when generating summaries, and models' efficiency in processing long inputs and outputs.  

\begin{figure*}[t]
\centering
\includegraphics[width=5.6in]{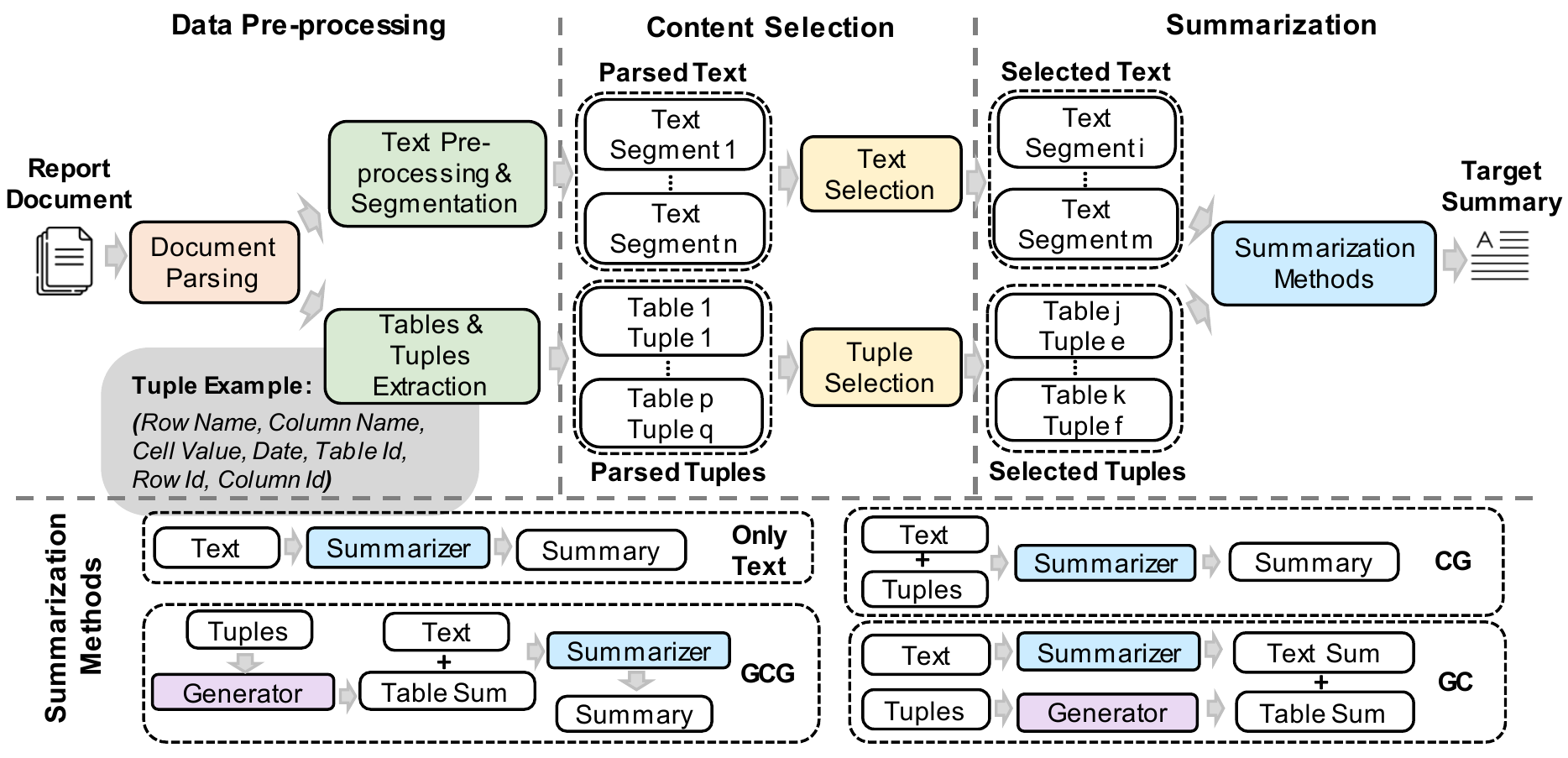}
\caption{An overview of our solution for long text and multi-table summarization.}\label{fig:FindSum_framework} 
\end{figure*}

Previous document summarization datasets usually regard non-textual content as noises and filter them out. When target summaries only focus on narratives and qualitative descriptions, removing non-textual content has little effect since the document's text already contains most of the required information.
However, when it comes to report documents, like financial reports, their summaries should cover both the narrative content and quantitative descriptions of critical metrics recorded in tables, which are essential for readers' analysis and decision-making \cite{sechowtoread}. Existing datasets cannot meet the requirements of summarizing long text and multiple tables in each report document.

To deal with the scarcity of available data, we propose FINDSum, the first large-scale dataset for long text and multi-table summarization.
FINDSum has two subsets named FINDSum-ROO and FINDSum-Liquidity for summarizing companies' results of operations and liquidity. 
As shown in Table \ref{table:stats_dataset_overall}, each example's inputs include tens of thousands of words and dozens of tables from a report document.
Besides, FINDSum's target summaries usually contain more numerical values than previous datasets. Meanwhile, most numerical values in target summaries cannot be found in the corresponding input text. Only focusing on text is not enough for summarizing financial reports.

Figure \ref{fig:FindSum_framework} shows our solution for long text and multi-table summarization. It has three main steps: data pre-processing, content selection, and summarization. 
The content selection step aims to compress long inputs while maximizing the recall of salient content in long text and dozens of tables.
Specifically, we adopt the Maximum Marginal Recall Gain (MMRG) method to select salient text segments as a part of inputs. 
As for the tabular content, we transform each table cell into a tuple and regard the salient tuple selection as a binary classification problem.
The summarization step should jointly consider the text and tabular data. 
We present three types of summarization methods: generate-and-combine (GC), combine-and-generate (CG), and generate-combine-and-generate (GCG). 

The complexity of the transformer's self-attention mechanism scales quadratically with the input length \cite{vaswani2017attention}. It can limit transformer-based models' efficiency.
Thus we employ content selection methods and sparse attention mechanisms to reduce the complexity and enable finetuning large pre-trained models over long inputs on an off-the-shelf GPU. 
Besides, existing autoregressive models still have difficulty in generating long sequences \cite{ranzato2016sequence,holtzman2019curious}. We employ a divide-and-conquer-based approach to generate summary segments in parallel and combine them as the final summary.

We benchmark advanced extractive and abstractive summarizers as baselines on our FINDSum dataset. To compare their performance, we conduct automatic evaluation and human evaluation. In addition to the commonly used ROUGE scores \cite{lin2004rouge}, we propose a set of evaluation metrics to assess the usage of numerical information in produced summaries.
Experimental results show that our methods can outperform competitive baselines.

\begin{table*}[t]
\small
\renewcommand\arraystretch{0.9}
\centering
\begin{tabular}{lccccccccc}
\toprule \textbf{Dataset} & \textbf{Pairs} & \textbf{\makecell*[c]{Words\\ (Doc)}} & \textbf{\makecell*[c]{Sents\\ (Doc)}}& \textbf{\makecell*[c]{Words\\ (Sum)}} & \textbf{\makecell*[c]{Sents\\ (Sum)}} & \textbf{\makecell*[c]{Num\\ (Sum)}} & \textbf{\makecell*[c]{\% Covered\\ Num}} & \textbf{Cov.} & \textbf{Dens.}\\
\midrule
CNN/DM & 312,085 & 810.6 & 39.8 & 56.2 & 3.7 & 0.6 & 78.7 & 0.9 & 3.8\\
PubMed & 133,215 & 3049.0 & 87.5 & 202.4 & 6.8 & 3.3 & 68.2 & 0.8 & 5.8\\
arXiv & 215,913 & 6029.9 & 205.7 & 272.7 & 9.6 & 0.7 & 53.9 & 0.9 & 3.8  \\
\midrule
FINDSum-ROO & 21,125 & 45,566.0 & 1250.5 & 660.7 & 16.3 & \textbf{24.3} & \textbf{26.3} & 0.9 & 9.7 \\
FINDSum-Liquidity & 21,125 & 45,566.0 & 1250.5 & 1,057.6 & 26.7 & \textbf{32.3} & \textbf{41.2} & 0.9 & 9.6\\ 
\bottomrule
\end{tabular}
\caption{Statistical information of summarization datasets. "Pairs" is the number of examples. "Words" and "Sents" denote the average number of words and sentences in input text or target summary. "Num" is the average number of numerical values in target summaries, and "Covered Num" is the ratio of the target summary's numerical values found in the input text. "Cov." and "Dens." are the extractive fragment's coverage and density.} 
\label{table:stats_dataset_overall}
\end{table*}

Our contribution is threefold:
\begin{itemize}
\item Our primary contribution is building FINDSum, the first large-scale dataset for long text and multi-table summarization.
\item We present three types of methods incorporating text and tables into summary generation.
\item We propose evaluation metrics to assess the usage of numerical information in summaries.
\end{itemize}

\section{Related Work}
\subsection{Automatic Document Summarization}
Automatic document summarization techniques can produce a concise summary covering the salient information within the input document.
In recent years, both large-scale summarization datasets and advanced neural models boosted the improvements in the quality of produced summaries. 
Expect for the widely studied news summarization \cite{grusky-etal-2018-newsroom,fabbri2019multi}, summarizing long documents received more attention in recent years. There are some datasets collected from different domains, including scientific literature \cite{cohan2018discourse,ijcai2022p591}, government reports \cite{huang-etal-2021-efficient}, and books \cite{kryscinski2021booksum}.
The Financial Narrative Summarisation shared task in 2020 \cite{el-haj-etal-2020-financial} delivered an annual report dataset from firms listed on the London Stock Exchange. These datasets only focus on the text, regard tabular data as noises, and filter them out.

Previous summarization methods can be generally classified into two categories: extractive \cite{erkan2004lexrank,mihalcea2004textrank} and abstractive \cite{Nallapati2016AbstractiveTS,zhang2020pegasus,liu2022key} summarization methods. 
To model longer input sequences with limited GPU memory, \citet{huang-etal-2021-efficient} compare various efficient attention mechanisms for the encoder and propose an encoder-decoder attention named Hepos. 
\citet{ijcai2022p591} identify and encode salient content in different aspects from diverse and long input content by category-based alignment and sparse attention mechanisms.
\citet{zhang2022summn} divide the summarization process into multiple stages and keep segmenting, summarizing, and concatenating long inputs till they are compressed to a fixed length. 
\citet{mao2022dyle} adopt the extract-then-generate method and jointly train the extractor and generator by combining loss functions.
Except for these summarization methods only focusing on the text,  \citet{jangra2021survey} comprehensively review the multi-modal summarization but still neglect tabular data in documents.

\subsection{Table Summarization}
There are some table summarization or table-to-text generation datasets, like the WEATHERGOV \cite{liang-etal-2009-learning}, WikiBio \cite{lebret-etal-2016-neural}, ROTOWIRE \cite{wiseman2017challenges}, and SBNATION \cite{wiseman2017challenges}. 
Some advanced methods, like hierarchical-encoder \cite{rebuffel2020hierarchical}, macro-plan \cite{puduppully-lapata-2021-data}, and LATTICE \cite{wang-etal-2022-robust}, achieved good performance on these datasets.
However, existing datasets and methods are usually limited to generating short descriptions for limited cells in a few tables with similar schemas. 
Conversely, each financial report usually contains numerous cells in dozens of different shaped tables. Selecting salient ones from thousands of cells can be challenging. 
In addition to summarizing multiple tables, we observe that human-written summaries can combine the information from both the text and multiple tables within the report document. 
Unstructured text and structured tabular data have different natures. It is also challenging to effectively integrate different types of input data when generating summaries.
To fill in the gap between existing datasets' limitations and the actual requirement of long text and multi-table summarization, we propose the FINDSum dataset and three types of summarization methods.

\section{FINDSum Dataset}

Financial report document summarization (FINDSum) is the first large-scale dataset for long text and multi-table summarization\footnote{Our dataset: https://github.com/StevenLau6/FINDSum}.
This section introduces our data collection and pre-processing procedures and describes FINDSum's two subsets. 
We conduct descriptive statistics and in-depth analysis on our FINDSum dataset and compare it with existing summarization datasets.

\begin{table}[t]
\small
\renewcommand\arraystretch{1.0}
\centering
\setlength{\tabcolsep}{1.2mm}{
\begin{tabular}{lcccc}
\toprule \textbf{\multirow{2}*{Dataset}} & \multicolumn{4}{c}{\textbf{\% of novel n-grams in target summary}} \\
~ & \textbf{unigrams} & \textbf{bigrams} & \textbf{trigrams} & \textbf{4-grams} \\
\midrule
CNN/DM & 19.50 & 56.88 & 74.41 & 82.83\\
PubMed & 18.38 & 49.97 & 69.21 & 78.42\\
arXiv & 15.04 & 48.21 & 71.66 & 83.26\\
\midrule
\makecell{FINDSum-\\ROO} & 17.79 & 50.59 & 72.13 & 81.66\\
\makecell{FINDSum-\\Liquidity} & 26.45 & 59.63 & 80.43 & 88.48 \\
\bottomrule
\end{tabular}}
\caption{The proportion of novel n-grams in target summaries of different summarization datasets.}
\label{table:stats_dataset_novel_ngrams}
\end{table}

\begin{figure}[t]
\centering
\includegraphics[width=3.0in]{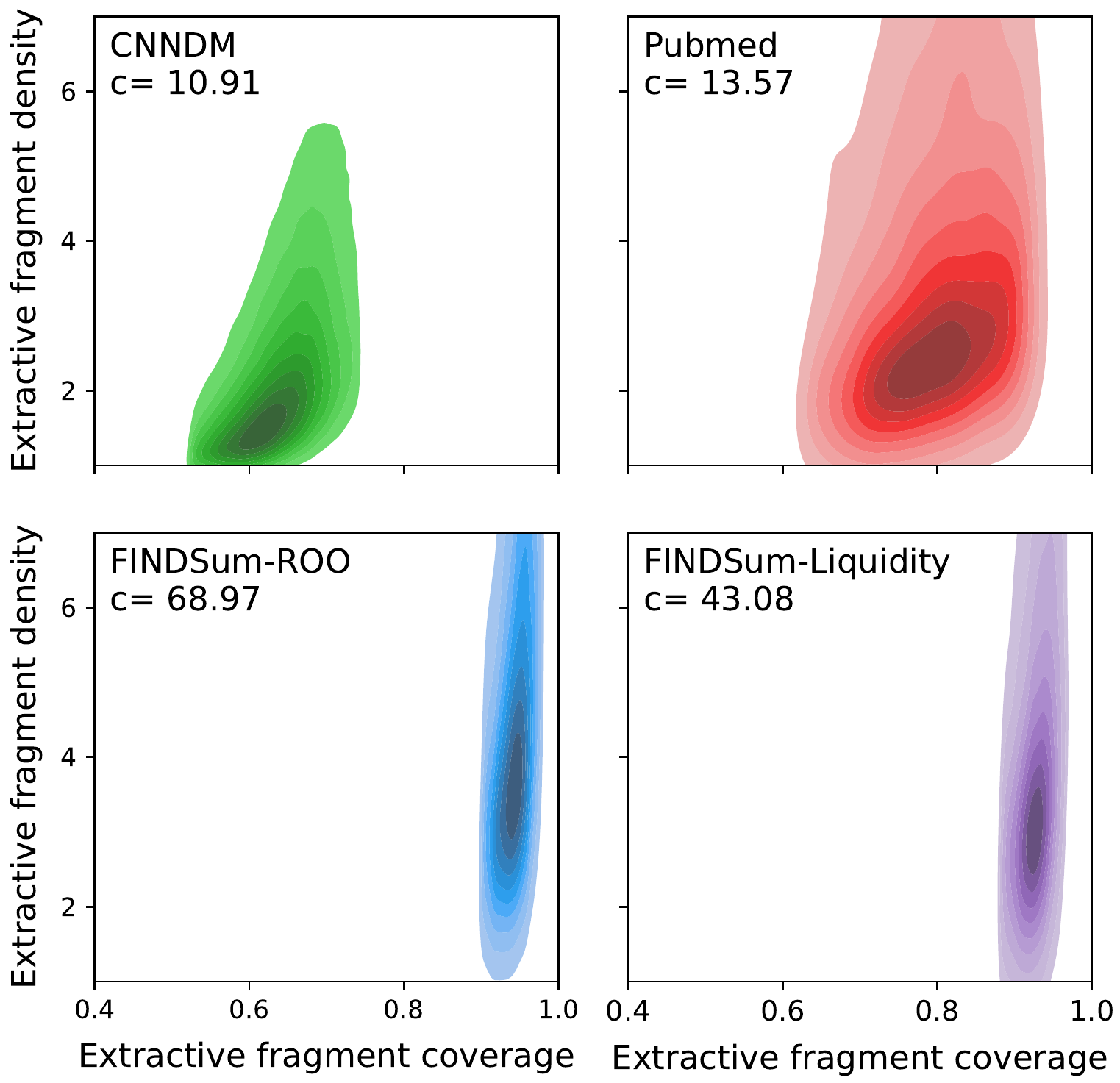}
\caption{Distributions of extractive fragment density and extractive fragment coverage.}\label{fig:diversity_visual} 
\end{figure} 

\subsection{Data Collection and Pre-processing}
\label{subsec:data_collect_preprocess}

Form 10-K is the annual report that comprehensively describes a company's financial performance in the prior fiscal year \cite{sechowtoread}.
We collected HTML files of 10-K forms from the Electronic Data Gathering, Analysis, and Retrieval (EDGAR) system \footnote{www.sec.gov/edgar/searchedgar/companysearch.html}. 
The U.S. Securities and Exchange Commission (SEC) makes companies' 10-K forms available to the public through the EDGAR system.
The SEC stipulates the 10-K form's format and required content. It usually contains four parts and sixteen items \cite{fotm10kgeneralinstruct}. 
The item "Management's Discussion and Analysis of Financial Condition and Results of Operations" (MD\&A) contains the management's summary of the company's results of operations and liquidity \cite{regsk303}. 
Our FINDSum dataset uses the text in MD\&A's two sections: "results of operations" and "liquidity and capital resources" as target summaries and the rest content of each report document as the input.

After collecting tens of thousands of 10-K forms' HTML files, we parse them and split each item's text and tables.
To align tables and text and keep tables' positional information, we add a special token containing each table's index into the table's original position in the document's text. 
Extracted text and tables are stored in separate files. 
Text and tabular data require different pre-processing procedures, considering their different natures.

Our text pre-processing procedures include: removing noises (e.g., cover pages before the first item and special characters composing a style) and dividing text in different parts of 10-k form into text segments. To pre-process tabular data, we need to extract table content (e.g., names of rows and columns, cell content), remove noises in table content, and transform each cell into a tuple: (row name, column name, cell value, date, table id, row id, column id). The cell value in the tuple concatenates the original cell value and the rounding result with an ampersand. 
Besides, we remove duplicate samples and outliers with too-short input text, truncate too-long input text, split the training (80\%), validation (10\%), and test (10\%) sets. Considering that the same company's annual reports in different years usually have duplicate content, we split these three sets by company to minimize their overlaps.

\subsection{Dataset Description}
We built the FINDSum dataset based on collected report documents. FINDSum has two subsets, which will be introduced in this subsection.

\noindent\textbf{FINDSum-ROO} is the subset focusing on each company's results of operations (ROO). 
In the "results of operations" section of MD\&A, the company's management usually compares and explains critical items of revenue and expense in the current and prior period \cite{fotm10kgeneralinstruct}. This section's text can be regarded as the target summary written by the expert. 
Table \ref{table:stats_dataset_overall} exhibits that the average number of numerical values in FINDSum-ROO's target summaries is dozens of times larger than that of previous datasets. However, nearly three-quarters of these numerical values cannot be found in the rest text of each report. 
A lot of critical numerical information is only recorded in tables. Therefore, we use both the rest parts' text and all the tables in each report document as inputs for each example.

\noindent\textbf{FINDSum-Liquidity} is built for summarizing each company's liquidity and capital resources. The "liquidity and capital resources" section in MD\&A mainly analyzes the company's ability to generate and obtain cash and its plans for cash \cite{regsk303}. This section's text can be used as the target summary in FINDSum-Liquidity. 
Similarly, most numerical values in target summaries are not included in the rest parts' text.
FINDSum-Liquidity's inputs also include the rest text and all the tables in each report.

\subsection{Dataset Analysis}

We conduct statistics and analysis on FINDSum's two subsets. Table \ref{table:stats_dataset_overall} shows that both the input documents and target summaries of these two subsets are much longer than those of previous summarization datasets. These two subsets' target summaries contain much more numerical information, while most of them cannot be found in the input text.

To measure how abstractive our target summaries are, we count the percentage of summaries' novel n-grams not appearing in inputs. 
Table \ref{table:stats_dataset_novel_ngrams} shows that FINDSum-Liquidity has a larger ratio of novel n-grams in target summaries, compared with other datasets. It reflects that the FINDSum-Liquidity is more abstractive, while the FINDSum-ROO's abstractiveness is similar to that of existing datasets. 
Besides, we calculate the coverage and density of extractive fragment \cite{grusky-etal-2018-newsroom} to assess these datasets' extractive nature. Table \ref{table:stats_dataset_overall} shows that the extractive fragment density of our dataset is higher than that of previous summarization datasets, while their extractive fragment coverage is similar. 
We also visualize the distributions of coverage and density by the kernel density estimation in Figure \ref{fig:diversity_visual}. 
The variability along the y-axis (density) suggests the varying writing styles of target summaries in our FINDSum dataset.

\section{Method}
\label{sec:model}
Summarizing long text and multiple tables has several challenging issues:
identifying the salient information from a large amount of input content, incorporating the text and tabular content into the summary generation, and efficiently processing long input and output sequences. This section presents our solution to the above issues.

\subsection{Content Selection}
\label{subsec:content_selection}
As shown in Figure \ref{fig:FindSum_framework}, our solution has three steps: data pre-processing, content selection, and summarization. 
After the pre-processing step, we can get dozens of text segments and thousands of tuples from dozens of tables in each report document.
The salient content usually scatters in text and tables, making it challenging to select the salient content accurately. 
We add the content selection step as a rough selection to compress long inputs while maximizing the recall of salient content that should be preserved in summaries. Then the compressed inputs are fed into the summarizer for further selection. Content selection methods' output length should not exceed a fixed length, as summarization models' complexity can scale with its input length.

We employ separate methods to select salient content from textual and tabular data considering their different natures.
To select salient text segments, we adopt a method named Maximum Marginal Recall Gain (MMRG) on our training set. 
Specifically, MMRG keeps adding the text segment bringing the maximum gain of n-gram's recall into the combination of selected segments till reaching the length limit. Finally, we can get selected salient segments' ids and choose text segments with the same ids for samples in our test set. 
MMRG's pseudocode is in Appendix \ref{appendixsec:content_selection}. 
We also follow \citet{liu2018generating} to try extractive summarizers, like Textrank \cite{mihalcea2004textrank} and Lexrank \cite{erkan2004lexrank}, for salient text selection. We use the recall of n-grams to evaluate these methods' performance in selecting the salient text of the same length. 
Table \ref{table:text_segment_selection} in Appendix \ref{appendixsec:content_selection} indicates that MMRG outperforms these extractive summarizers, so we use it for text segment selection.

As for those thousands of tuples extracted from tables, we regard the salient tuple selection as a binary classification problem. Based on the FINDSum dataset, we annotate a tuple selection dataset for training and evaluating different classification methods (e.g., logistic regression, support vector machine, Adaboost \cite{hastie2009multi}, and XGBoost \cite{chen2016xgboost})\footnote{We use the implementation of XGBoost from xgboost.readthedocs.io/en/stable/ and other classifiers from scikit-learn.org/stable/}. 
We also try utilizing various features, including positional features (e.g., indexes of the row, column, table, and section, together with their normalized values) and text features (e.g., word embedding or one-hot keywords representation of row name and column name). Considering the content selection step focuses more on the recall of salient content, we sort these tuples by their positive probability predicted by the trained classifier and use the top-n tuples' recall to evaluate these classifiers. Table \ref{table:tuple_selection} in Appendix \ref{appendixsec:content_selection} shows evaluation results. The XGBoost model equipped with positional features and Glove embedding \cite{pennington2014glove} outperforms other classifiers, so we use it for tuple selection.

\subsection{Summarizing Textual and Tabular Data}
\label{subsec:TTS_methods}

To incorporate text and tabular data into summary generation, we present three types of methods: generate-and-combine (GC), combine-and-generate (CG), and generate-combine-and-generate (GCG). We show their structures in Figure \ref{fig:FindSum_framework} and introduce them in this subsection.

\noindent\textbf{GC} treats the long text and multi-table summarization as two parallel processes. It assigns the maximum output lengths for the text summary and table summary, generates these two summaries separately, and concatenates them to form the final summary. GC has obvious limitations: 1) It cannot merge the information from text and tables when generating each summary sentence. 2) The pre-defined length assignment is not flexible enough to adapt to diverse examples.

\noindent\textbf{CG} first concatenates the selected text segments and tuples with a special symbol and then feeds them into a sequence-to-sequence summarizer. It requires the summarizer to learn text-to-text and tuple-to-text generation and jointly consider these two types of input content when generating summaries. Considering the selected tuples are from different tables whose shapes differ greatly, we only keep the first four items of each input tuple and leave out the ids of the row, column, and table.

\noindent\textbf{GCG} employs a tuple-to-text generator to produce input tuples' text descriptions. It concatenates the input text with the generator's output text and feeds them into the summarizer. Compared with the CG, GCG simplifies the requirement on the summarizer to focus on summarizing text, but the extra tuple-to-text generation process can lose some tuples' information.
We annotate a tuple-to-text generation dataset based on our FINDSum dataset for training and evaluating various generators.
Table \ref{table:eval_tuple2text_generation} indicates that the BART-large outperforms other baselines, so we use it as the tuple-to-text generator.

\subsection{Dealing with Long Inputs and Outputs}
Input documents in our FINDSum-ROO and FINDSum-Liquidity subsets contain tens of thousands of words. The average length of target summaries in FINDSum-Liquidity exceeds 1,000 words.
Long inputs and outputs bring some problems: 
1) The transformer model's self-attention mechanism \cite{vaswani2017attention} scales quadratically with the length of the input sequence, which is prohibitively expensive for long input \cite{choromanski2020rethinking} and precludes the usage of large pre-trained models with limited computational resources.
2) Existing autoregressive abstractive summarization methods still have difficulty in generating long text in terms of efficiency and quality \cite{ranzato2016sequence,holtzman2019curious}.
To deal with the first problem, we employ sparse attention mechanisms \cite{zaheer2020big,beltagy2020longformer} in our summarization models' encoders. The content selection step in our solution also reduces the length of input sequences. To handle the second problem, we follow a divide-and-conquer method \cite{gidiotis2020divide} and decompose the long summary generation problem into multiple sub-problems of summary segment generation. These summary segments can be generated in parallel and combined as a final summary. 
To minimize output summaries' redundancy, we add a constraint that the MMRG in the content selection step should not select the same combination of input text segments for generating different summary segments.

\subsection{Evaluation Metrics}
\label{subsec:evalmetrics}

We propose a set of evaluation metrics to assess the usage of numerical information in produced summaries. It is necessary for long text and multi-table summarization. 
We use $D$, $S$, and $H$ to denote the input document, human-written target summary, and the summarizer's output summary. $D_n$, $S_n$, and $H_n$ are sets of numbers contained in them. $|D_n|$, $|S_n|$, $|H_n|$ denote the sizes of these number sets.
For a produced summary $H$, we first extract the number set $H_{n}$ from it.\footnote{We do not count numbers in a word, like COVID-19.}  
Then $M(H_n,S_n)$ counts numbers in both the produced summary $H$ and the target summary $S$. $M(D_n,S_n)$ counts numbers appearing in both the input document $D$ and the target summary $S$.

We mainly consider three metrics: Number Precision (NP), Number Coverage (NC), and Number Selection (NS).
Calculated by Equation (\ref{equa:NP}), NP is the ratio of numbers in the produced summary that also appears in the target summary. It measures how well the produced summary matches the target summary in terms of contained numbers.
NC measures how well the produced summary covers the numbers appearing in both the target summary and the input document. 
Some of the numbers in the target summary cannot be directly found in the inputs (including textual and tabular data) and need numerical reasoning. Some of them may be lost when preparing the summarization model's inputs, which can limit produced summary's number recall computed by Equation (\ref{equa:NR}). 
To evaluate the summarization model's coverage capability, we divide the produced summary's number recall by the input document's number recall in Equation (\ref{equa:NC1}).
NS calculates the harmonic mean of NP and NC in Equation (\ref{equa:NS}) and reflects the quality of number selection in the produced summary.

\begin{equation} 
\mathrm{NP}(H_n,S_n)\!=\!\frac{M(H_n,S_n)}{|H_n|}
\label{equa:NP}
\end{equation}
\begin{subequations} 
\begin{align}
\mathrm{NR}(H_n,S_n)\!=&\frac{M(H_n,S_n)}{|S_n|}\label{equa:NR}\\
\mathrm{NC}(D_n,H_n,S_n)\!=&\frac{\mathrm{NR}(H_n,S_n)\!*|S_n|}{M(D_n,S_n)}\label{equa:NC1}
\end{align}
\end{subequations}
\begin{equation} 
\mathrm{NS}(D_n,H_n,S_n)\!=\!\frac{2*\mathrm{NP}*\mathrm{NC}}{\mathrm{NP}+\mathrm{NC}}
\label{equa:NS}
\end{equation}

\section{Experiments}
\label{sec:experiments}

\begin{table*}[t]
\renewcommand\arraystretch{1.1}
\small
\centering
\setlength{\tabcolsep}{1.5mm}{
\begin{tabular}{lccccccc|ccccccccc}
\hline
\textbf{\multirow{2}*{Type}}&\textbf{\multirow{2}*{Method}} & \multicolumn{6}{c|}{\textbf{FINDSum-Liquidity}} & \multicolumn{6}{c}{\textbf{FINDSum-ROO}}\\
\cline{3-8} \cline{9-14}
~ & ~ & \textbf{R-1} & \textbf{R-2} & \textbf{R-L} & \textbf{NP} & \textbf{NC} & \textbf{NS} & \textbf{R-1} & \textbf{R-2} & \textbf{R-L} & \textbf{NP} & \textbf{NC} & \textbf{NS} \\
\hline
\multirow{6}*{\makecell{Only\\Text}} & LexRank & 40.67 & 10.61 & 16.28 & 12.58 & 14.50 & 13.47 & 34.43 & 7.73 & 14.92 & 14.77 & 9.73 & 11.73 \\
 & TextRank & 41.71 & 10.90 & 16.54 & 13.37 & 13.02 & 13.19 & 35.93 & 7.74 & 15.08 & 14.68 & 10.96 & 12.55 \\
 & BART & 52.37 & 17.91 & 19.59 & 21.18 & 22.78 & 21.95 & 49.00 & 16.88 & 19.14 & 14.38 & 23.72 & 17.91 \\
 & PEGASUS & 52.57 & 18.46 & 19.75 & 16.98 & 22.74 & 19.44 & 51.92 & 19.31 & 21.47 & 10.90 & 21.89 & 14.55 \\
 & LED & 53.52 & 18.91 & 19.75 & 18.68 & 22.56 & 20.44 & 53.06 & 20.33 & 22.28 & 14.25 & 22.99 & 17.59\\
 & \makecell{BigBird-\\PEGASUS} & 53.42 & 19.39 & 20.07 & 17.16 & 22.44 & 19.45 & 53.08 & 20.85 & 20.94 & 13.15 & 23.82 & 16.95 \\
\hline
\multirow{2}*{\makecell{GC}} & $\mathrm{GC_{LED}}$ & 52.30 & 20.09 & 19.61 & 15.13 & \textbf{44.47} & 22.58 & 53.19 & 21.97 & 22.84 & 12.83 & \textbf{41.54} & 19.60 \\
 & $\mathrm{GC_{BigBird}}$ & 51.61 & 20.00 & 19.86 & 14.76 & 44.21 & 22.13 & 53.13 & 22.03 & 23.11 & 12.49 & 41.30 & 19.18\\
\hline
\multirow{2}*{\makecell{CG}} & $\mathrm{CG_{LED}}$ & 54.12 & 20.26 & 20.46 & \textbf{21.86} & 35.14 & \textbf{26.95} & 54.24 & 22.08 & 23.10 & 16.41 & 33.89 & 22.11\\
 & $\mathrm{CG_{BigBird}}$ & 53.82 & 20.15 & 20.39 & 20.98 & 34.29 & 26.03 & \textbf{54.40} & \textbf{22.48} & \textbf{23.21} & \textbf{16.46} & 35.84 & \textbf{22.56} \\
\hline
\multirow{2}*{\makecell{GCG}} & $\mathrm{GCG_{LED}}$ & \textbf{54.55} & 20.36 & 20.41 & 21.15 & 34.52 & 26.23 & 54.32 & 21.92 & 23.03 & 16.03 & 32.54 & 21.48\\
& $\mathrm{GCG_{BigBird}}$ & 53.90 & \textbf{20.47} & \textbf{20.59} & 20.67 & 36.43 & 26.38 & 54.12 & 22.11 & 23.02 & 15.33 & 32.82 & 20.90 \\
\hline
\end{tabular}}
\caption{\label{autoeval:combined_summary} Automatic evaluation results on test sets of FINDSum-Liquidity and FINDSum-ROO.}
\end{table*}

\begin{table}[t]
\small
\renewcommand\arraystretch{1.0}
\centering
\setlength{\tabcolsep}{1.2mm}{
\begin{tabular}{lcccc}
\toprule 
 & \textbf{R-1} & \textbf{R-2} & \textbf{R-L} & \textbf{BLEU}\\
\midrule
\multicolumn{2}{l}{\textbf{ROO}} & & &\\
T5-base & 45.45 & 24.77 & 28.84 & 12.20 \\
T5-large & 45.81 & 24.64 & 29.04 & 12.87 \\
BART-base &  42.08 & 20.45 & 25.86 & 10.57 \\
BART-large & \textbf{47.21} & \textbf{25.63} & \textbf{31.08} & \textbf{13.14} \\
\midrule
\multicolumn{2}{l}{\textbf{Liquidity}} & & &\\
T5-base & 48.90 & \textbf{28.34} & 31.98 & 15.44\\
T5-large & 49.03 & 28.05 & 32.02 & 15.86\\
BART-base & 45.71 & 24.75 & 29.28 & 13.66 \\
BART-large & \textbf{49.78} & 28.24 & \textbf{32.59} & \textbf{16.05} \\
\bottomrule
\end{tabular}}
\caption{Evaluation results of tuple-to-text generation.}
\label{table:eval_tuple2text_generation}
\end{table}

\subsection{Baselines}
In our experiments, we adopt advanced extractive and abstractive summarization models as baselines only using input text.

\noindent\textbf{LexRank and TextRank} \cite{erkan2004lexrank,mihalcea2004textrank} are two graph-based ranking methods that can be used for unsupervised extractive summarization. 

\noindent\textbf{BART} \cite{lewis2020bart} is a denoising autoencoder built with a sequence-to-sequence model that is pre-trained to reconstruct the original input text from the corrupted text. 

\noindent\textbf{PEGASUS} \cite{zhang2020pegasus} is a transformer-based model pre-trained with the Gap Sentences Generation (GSG) and Masked Language Model (MLM) objectives.

\noindent\textbf{T5} \cite{raffel2020exploring} is an encoder-decoder model pre-trained on a mixture of multiple unsupervised and supervised tasks.

\noindent\textbf{BigBird-PEGASUS} \cite{zaheer2020big} adopts the BigBird encoder with sparse attention mechanisms and the PEGASUS decoder. 

\noindent\textbf{Longformer-Encoder-Decoder (LED)} \cite{beltagy2020longformer}
follows BART’s architecture and adopts sparse attention mechanisms in its encoder.

\subsection{Experimental Setting}

\begin{table*}[t]
\renewcommand\arraystretch{1.1}
\small
\centering
\setlength{\tabcolsep}{1.7mm}{
\begin{tabular}{lccccccc|ccccccccc} 
\hline
\textbf{\multirow{2}*{\makecell{Text/\\Tuple}}}&\textbf{\multirow{2}*{Method}} & \multicolumn{6}{c|}{\textbf{FINDSum-Liquidity}} & \multicolumn{6}{c}{\textbf{FINDSum-ROO}}\\
\cline{3-8} \cline{9-14}
~ & ~ & \textbf{R-1} & \textbf{R-2} & \textbf{R-L} & \textbf{NP} & \textbf{NC} & \textbf{NS} & \textbf{R-1} & \textbf{R-2} & \textbf{R-L} & \textbf{NP} & \textbf{NC} & \textbf{NS} \\

\hline
\multirow{2}*{\makecell{1:1}} & $\mathrm{GC_{LED}}$ & 52.30 & 20.09 & 19.61 & 15.13 & \textbf{44.47} & \textbf{22.58} & 53.19 & 21.97 & 22.84 & 12.83 & \textbf{41.54} & 19.60 \\
 & $\mathrm{GC_{BigBird}}$ & 51.61 & 20.00 & \textbf{19.86} & 14.76 & 44.21 & 22.13 & 53.13 & 22.03 & \textbf{23.11} & 12.49 & 41.30 & 19.18\\
\hline
\multirow{2}*{\makecell{2:1}} & $\mathrm{GC_{LED}}$ & 52.28 & 18.37 & 19.12 & \textbf{16.63} & 22.45 & 19.11 & 53.56 & 21.95 & 22.78 & 13.45 & 36.54 & 19.66 \\
& $\mathrm{GC_{BigBird}}$ & 52.99 & \textbf{20.18} & 19.81 & 14.43 & 35.62 & 20.54 & 53.51 & 22.02 & 22.69 & 12.82 & 38.74 & 19.26\\
\hline
\multirow{2}*{\makecell{3:1}} & $\mathrm{GC_{LED}}$ & 52.57 & 18.47 & 19.21 & 16.13 & 22.24 & 18.70 & \textbf{53.66} & 21.88 & 22.48 & \textbf{13.62} & 36.21 & \textbf{19.79}\\
& $\mathrm{GC_{BigBird}}$ & \textbf{53.33} & 20.15 & 19.81 &  14.58 & 32.30 & 20.09 & 53.59 & \textbf{22.07} & 22.73& 13.18 & 35.84 & 19.27\\
\hline
\end{tabular}}
\caption{\label{autoeval:gc_diff_ratio} GC methods' automatic evaluation results on test sets of FINDSum-Liquidity and FINDSum-ROO. "Text/Tuple" denotes the assigned length ratio between the text and table summary in each combined summary.}
\end{table*}

\begin{table}[t]
\small
\renewcommand\arraystretch{1.0}
\centering
\begin{tabular}{lcccc}
\toprule & \textbf{Win} & \textbf{Lose} & \textbf{Tie} & \textbf{Kappa}\\
\midrule
\multicolumn{4}{l}{\textbf{FINDSum-ROO}}\\
Informativeness & 43.3\% & 20.8\%& 35.8\%& 0.653\\
Fluency & 27.5\%& 24.2\%& 48.3\%& 0.613\\
Non-Redundancy & 33.3\%& 21.7\%& 45.0\%& 0.644\\
\midrule
\multicolumn{4}{l}{\textbf{FINDSum-Liquidity}}\\
Informativeness & 41.7\% & 21.6\%& 36.7\%& 0.655\\
Fluency & 25.8\% & 25.0\%& 49.2\%& 0.611\\
Non-Redundancy & 32.5\% & 23.3\%& 44.2\%& 0.638\\
\bottomrule
\end{tabular}
\caption{Human evaluation results. “Win” represents the generated summary of our GCG-BigBird method is better than that of BigBird-PEGASUS in one aspect.}
\label{humaneval:results_GCG}
\end{table}

The vocabulary's maximum size is 50,265 for these abstractive summarization models.
When finetuning these pre-trained models on our datasets, we use the learning rate of $5e^{-5}$, and adopt the learning rate warmup and decay. The optimizer is Adam with $\beta_1$=0.9 and $\beta_2$=0.999.
We use dropout with the probability 0.1. 
In the generation process, we use beam search with a beam size of 5. Trigram blocking is used to reduce repetitions. 
We adopt the implementations of BART, PEGASUS, BigBird, and LED from HuggingFace's Transformers \cite{wolf2020transformers}. All the models are trained on one NVIDIA RTX 8000 GPU.

\subsection{Results and Discussion}
\label{subsec:results_and_discuss}

We present and analyze our experimental results in this subsection. 
To compare the quality of summaries produced by different models, we conduct automatic and human evaluations. We also perform the ablation study to validate the effectiveness of components in our methods. Output examples of different summarization models and tuple-to-text generators are presented in Appendix \ref{appendixsec:output_example}.

\begin{table}[t]
\small
\renewcommand\arraystretch{1.0}
\centering
\begin{tabular}{lcccc}
\toprule & \textbf{Win} & \textbf{Lose} & \textbf{Tie} & \textbf{Kappa}\\
\midrule
\multicolumn{4}{l}{\textbf{FINDSum-ROO}}\\
Informativeness & 44.2\% & 20.8\%& 35.0\%& 0.626\\
Fluency & 26.7\%& 25.8\%& 47.5\%& 0.616\\
Non-Redundancy & 35.0\%& 23.3\%& 41.7\%& 0.632\\
\midrule
\multicolumn{4}{l}{\textbf{FINDSum-Liquidity}}\\
Informativeness & 40.8\% & 20.8\%& 38.3\%& 0.620\\
Fluency & 25.0\% & 24.2\%& 50.8\%& 0.615\\
Non-Redundancy & 31.7\% & 22.5\%& 45.8\%& 0.626\\
\bottomrule
\end{tabular}
\caption{Human evaluation results. “Win” represents the generated summary of our CG-BigBird method is better than that of BigBird-PEGASUS in one aspect.}
\label{humaneval:results_CG}
\end{table}

In the automatic evaluation, we calculate the ROUGE $\mathrm{F}_1$ scores \cite{lin2004rouge}, including the overlaps of unigrams (R-1), bigrams (R-2), and longest common subsequence (R-L)\footnote{github.com/falcondai/pyrouge/}, and our NP, NC, and NS scores.
Table \ref{autoeval:combined_summary} reports the final combined summaries' scores. 
Each summary segment's ROUGE scores are exhibited in Table \ref{autoeval:summary_segment} of Appendix \ref{appendixsec:eval_results}. 
These abstractive summarizers based on pre-trained models outperform unsupervised extractive summarizers.
Besides, baselines equipped with sparse attention mechanisms \cite{zaheer2020big,beltagy2020longformer} can model longer context and achieve higher ROUGE scores.
Covering more salient content scattered in longer inputs can benefit output summaries' informativeness. 

Our CG and GCG methods outperform these text-only baselines on FINDSum's two subsets. 
Incorporating tabular information is conducive to improving the NP, NC, NS, and ROUGE scores. 
GCG methods perform better on FINDSum-Liquidity, while CG methods perform better on FINDSum-ROO. Table \ref{table:stats_dataset_overall} shows that target summaries in the FINDSum-ROO subset have a larger ratio of numerical information not found in the input text and rely more on tables. 
The table content passes one generation process in CG methods but needs to pass through two generation processes in GCG methods. The extra tuple-to-text generation can lose some required tabular information and accumulate more errors. In FINDSum-Liquidity, a larger ratio of the numerical information can be found in the input text, and the loss of tabular information in the extra generation process has less effect.

We evaluate multiple tuple-to-text generators by the ROUGE  \cite{lin2004rouge} and BLEU scores\footnote{www.nltk.org/api/nltk.translate.bleu\_score.html. We report the cumulative 4-gram BLEU score.} \cite{papineni2002bleu}.
Table \ref{table:eval_tuple2text_generation} depicts the performance of different tuple-to-text generators on ROO and Liquidity subsets. The large model of BART \cite{lewis2020bart} performs the best on these two subsets.
These generators perform better on the Liquidity subset. The better performance of the tuple-to-text generator also contributes to the GCG methods' performance on the FINDSum-Liquidity.  

These GC methods do not perform well, which is due to GC's limitations mentioned in subsection \ref{subsec:TTS_methods}. Table \ref{autoeval:combined_summary} shows the evaluation result of combined summaries, in which half of the content is text summary and the other half is table summary. Although they can achieve high NC scores, their NP and ROUGE scores are unsatisfactory.
The result reflects that it is not appropriate to treat long text and multi-table summarization as two parallel processes. The inflexible length assignment is difficult to set for diverse examples. We show its effect on generated summaries' quality in Table \ref{autoeval:gc_diff_ratio}.

\begin{table}[t]
\small
\renewcommand\arraystretch{1.0}
\centering
\begin{tabular}{lccc}
\toprule
 & \textbf{R-1} & \textbf{R-2} & \textbf{R-L}\\
\midrule
\textbf{FINDSum-ROO}\\
GCG-BigBird & \textbf{54.12} & \textbf{22.11} & \textbf{23.02}\\
w/o tabular data & 53.08 & 20.85 & 20.94\\
w/o sparse attn & 51.92 & 19.31 & 21.47\\
w/o input text & 47.19 & 17.89 & 21.09\\
\midrule
\textbf{FINDSum-Liquidity}\\
GCG-BigBird & \textbf{53.90} & \textbf{20.47} & \textbf{20.59}\\
w/o tabular data & 53.42 & 19.39 & 20.07\\
w/o sparse attn & 52.57 & 18.46 & 19.75\\
w/o input text & 44.17 & 15.60 & 18.49\\
\bottomrule
\end{tabular}
\caption{\label{table:evalablation} Ablation study on test sets of FINDSum-ROO and FINDSum-Liquidity.}
\end{table}

We performed the human evaluation to compare different models' output summaries in terms of informativeness (the coverage of information from input documents), fluency (content organization and grammatical correctness), and non-redundancy (less repetitive information). We randomly selected 30 samples from the test sets of the FINDSum-ROO and FINDSum-Liquidity subsets, respectively. Four annotators are required to compare two models' output summaries that are presented anonymously. We also assess their agreements by Fleiss' kappa \cite{fleiss1971measuring}. Table \ref{humaneval:results_GCG} and \ref{humaneval:results_CG} exhibit that GCG-BigBird and CG-BigBird significantly outperform the BigBird-PEGASUS only using input text in terms of informativeness and are comparable in terms of fluency and non-redundancy.

We also conduct the ablation study to validate the effectiveness of components in our GCG-BigBird method. 
In Table \ref{table:evalablation}, "w/o tabular data" refers to the BigBird-PEGASUS model only using input text. 
The results show that incorporating tabular data benefits the report document summarization. The sparse attention mechanisms in the encoder also benefit our model's performance.  
Besides, we tried only using the tuple-to-text generation result as the produced summary. "w/o input text" denotes the output of the BART-large-based tuple-to-text generator\footnote{As summaries, they are compared with target summaries instead of target outputs in the tuple-to-text generation.}. The performance degradation reveals that it is important to 
jointly consider input textual and tabular data in the report summary generation.

In the future, we intend to explore more methods and evaluation metrics for long text and multi-table summarization. There is still room to improve the produced summaries' quality and summarization methods' efficiency. Evaluation metrics assessing the produced summaries' factual correctness and fidelity to the input content are also necessary.
Long text and multi-table summarization is still an open problem, and there is still a lot of work to do.

\section{Conclusion}
In this paper, we introduce FINDSum, the first large-scale dataset for long text and multi-table summarization.
Built on tens of thousands of annual report documents from thousands of companies, FINDSum has two subsets for summarizing these companies' results of operations and liquidity. 
Besides, we propose a solution for the long text and multi-table summarization. It has three main steps: data pre-processing, content selection, and summarization. 
We adopt different content selection methods to select the salient content from the long text and dozens of tables in each report document. 
As for the summarization step, we present and compare three types of summarization methods incorporating text and tabular data into the summary generation.
To assess the usage of numerical information in produced summaries, we propose a set of evaluation metrics.
Dataset analyses and experimental results indicate the importance of jointly considering input textual and tabular data when generating summaries for report documents.

\section*{Limitations}
Our work still has some limitations. 
Although we adopt content selection methods, sparse attention mechanisms, and the divide-and-conquer-based training approach to enable finetuning large pre-trained models over long inputs and outputs on an off-the-shelf GPU, finetuning still needs tens of hours. These abstractive summarization models' efficiency needs further improvements. 
When observing generation results, we found current neural abstractive summarization and text generation models have flaws in the fidelity to the input content, which can bring hallucinations in output text \cite{zhao2020reducing}. It is a common problem in text generation research and needs further study.

\section*{Ethics Statement}
We build FINDSum based on the publicly available regular filings of listed companies. The U.S. Securities and Exchange Commission's EDGAR (Electronic Data Gathering, Analysis, and Retrieval) system provides these filings as public information that can be copied or distributed.
We follow EDGAR's guidelines on data accessing and collect data from its public APIs. 
Models trained with our dataset are primarily used to support humans in improving the efficiency of financial analysis instead of a substitute for human experts.

\section*{Acknowledgements}
This work was supported by the Research Institute for Artificial Intelligence of Things, The Hong Kong Polytechnic University, and the Hong Kong Jockey Club Charities Trust (Project S/N Ref.: 2021-0369).

\bibliography{anthology,custom}
\bibliographystyle{acl_natbib}

\appendix

\section{Appendix}
\label{sec:appendix}

\subsection{Content Selection Methods}
\label{appendixsec:content_selection}

\begin{table*}[t]
\renewcommand\arraystretch{1.0}
\small
\centering
\begin{tabular}{l|cc|cc|cc|cc|cc}
\hline
& \multicolumn{4}{c|}{\textbf{FINDSum-ROO}}& \multicolumn{6}{c}{\textbf{FINDSum-Liquidity}}\\
\cline{2-5} \cline{6-11}
\textbf{\multirow{2}*{Method}} & \multicolumn{2}{c|}{\textbf{Segment 1}} & \multicolumn{2}{c|}{\textbf{Segment 2}} &\multicolumn{2}{c|}{\textbf{Segment 1}} & \multicolumn{2}{c|}{\textbf{Segment 2}} & \multicolumn{2}{c}{\textbf{Segment 3}} \\
\cline{2-3} \cline{4-5} \cline{6-7} \cline{8-9} \cline{10-11}
~ & $\mathrm{Recall_1}$ & $\mathrm{Recall_{a}}$ & $\mathrm{Recall_1}$ & $\mathrm{Recall_{a}}$ & $\mathrm{Recall_1}$ & $\mathrm{Recall_{a}}$ & $\mathrm{Recall_1}$ & $\mathrm{Recall_{a}}$ & $\mathrm{Recall_1}$ & $\mathrm{Recall_{a}}$\\
\hline
LexRank & 56.01 & 22.14 & 53.96 & 20.72 & 49.71 & 18.59 & 48.92 & 17.97 & 46.45 & 17.00 \\
TextRank & 58.38 & 22.94 & 56.25 & 21.53 & 55.18 & 20.94 & 54.02 & 20.40 & 51.72 & 19.49 \\
MMRG & \textbf{63.38} & \textbf{28.01} & \textbf{61.68} & \textbf{27.85} & \textbf{58.61} & \textbf{24.28} & \textbf{56.69} & \textbf{23.09} & \textbf{53.94} & \textbf{21.62} \\
\hline
\end{tabular}
\caption{Evaluation results of input text selection methods. $\mathrm{Recall_1}$ denotes the recall of unigram, and $\mathrm{Recall_{a}}$ is the average recall of unigram, bigram, trigram, and 5-gram.}
\label{table:text_segment_selection} 
\end{table*}

As introduced in subsection \ref{subsec:content_selection}, the content selection step filters out the non-prominent content and retains the salient content as summarizers' inputs. We employ different methods to select salient content from text and tabular data, considering their different natures.
To select the salient text segments, we adopt a statistics-based method named Maximum Marginal Recall Gain (MMRG) on our training set. MMRG's outputs include selected salient segments' ids. 
Then we choose text segments with the same ids for samples in our test set.
Algorithm \ref{algorithm:marginal_recall_gain} is MMRG's pseudocode. 
We also try some extractive summarization methods, like textrank and lexrank. 
Table \ref{table:text_segment_selection} shows that MMRG outperforms these extractive summarizers, so we use it in the text segment selection.

\begin{algorithm} 
	\caption{\small{Maximum Marginal Recall Gain (MMRG)}} 
	\label{algorithm:marginal_recall_gain}
	\begin{algorithmic}
	\small
		\Require Input $m$ examples $I \gets [e_1,...,e_m]$, each example $e_i$ contains $n$ parts for selection $e_i \gets [p_i^1,...,p_i^n]$, the list of target item $T \gets [t_1,...,t_m]$, and the maximum number of selected parts $n^{\prime}$ ($n^{\prime} \ll n$)
		\Ensure The list of selected parts' id $S \gets [j,...,k]$ and the selected inputs $I^{\prime} \gets [e_1^{\prime},...,e_m^{\prime}]$, in which each example $e_i^{\prime}$ has selected parts $e_i^{\prime} \gets [p_i^j,...,p_i^k]$ $(|e_i^{\prime}|=|S|\leq n^{\prime})$
		\State 
		\Function{RecallGain}{$I,I^{\prime}, T, j$}
		    \State $i \gets 1$;
            \State $rgain_{sum} \gets 0$;
            \While{$i \leq m$}
                \State $p_i^j \gets I[i][j]$;
                \State $concat\_str \gets         \mathrm{Concat}(I^{\prime}[i],p_i^j)$;
                \State // Calculate the recall gain brought by the j-th part
                \State $rgain \gets         \mathrm{Recall}(concat\_str,T[i])-\mathrm{Recall}(I^{\prime}[i],T[i])$; 
                \State $rgain_{sum} \gets rgain_{sum}+rgain$;
                \State $i \gets i + 1$;
                \EndWhile
            \State $rgain_{avg} \gets rgain_{sum}/m$;
            \State \Return{$rgain_{avg}$}
            \EndFunction
        \State 
        \Function {SelectPart}{$I, I^{\prime}, T, S$}
            \State $j \gets 1$;
            \State $rgain_{max} \gets 0$;
            \State $j_{select} \gets 0$;
            \State //Find the part $p^{j_{select}}$ bringing the largest average recall gain
            \While{$j \leq n$}
                \If{$j$ not in $S$}
                    \State $rgain_{avg}=\mathrm{RecallGain}(I,I^{\prime}, T, j)$;
                    \If{$rgain_{avg} > rgain_{max}$}
                        \State $j_{select} \gets j$;
                        \State $rgain_{max} \gets rgain_{avg}$;
                        \EndIf
                \EndIf
                \State $j \gets j+1$;
                \EndWhile
            \State \Return{$j_{select}$;}
            \EndFunction
        \State 
        \Function {MMRG}{$I, T, n^{\prime}$}
            \State $S \gets [\;]$;
            \State $e_1^{\prime},...,e_m^{\prime} \gets ``",...,``"$;
            \State $I^{\prime} \gets [e_1^{\prime},...,e_m^{\prime}]$;
            \While{$|S| < n^{\prime}$}
                \State $j_{select}=\mathrm{SelectPart}(I, I^{\prime}, T, S)$;
                \If{$\mathrm{j_{select}} > 0$}
                    \State $S \gets S \cup [j_{select}]$;
                    \While{$i \leq m$}
                        \State $I^{\prime}[i] \gets \mathrm{Concat}(I^{\prime}[i],p_i^{j_{select}})$;
                        \EndWhile
                    \EndIf
                \EndWhile
                \State \Return{$S,I^{\prime}$;}
            \EndFunction

	\end{algorithmic} 
\end{algorithm}

\begin{table*}[t]
\small
\renewcommand\arraystretch{1.0}
\centering
\begin{tabular}{lcccccccccccc}
\toprule \textbf{\multirow{3}*{Method}} & \textbf{\multirow{3}*{Features}} & \multicolumn{5}{c}{\textbf{Liquidity}}& &\multicolumn{5}{c}{\textbf{ROO}} \\
\cline{3-7} \cline{9-13}
~ & ~ & \multicolumn{2}{c}{\textbf{Top-100}}& & \multicolumn{2}{c}{\textbf{Top-200}}& &\multicolumn{2}{c}{\textbf{Top-100}}& &\multicolumn{2}{c}{\textbf{Top-200}}\\
\cline{3-4} \cline{6-7} \cline{9-10} \cline{12-13}
~ & ~ & \textbf{ACC} & \textbf{Recall} & & \textbf{ACC} & \textbf{Recall} & & \textbf{ACC} & \textbf{Recall} & & \textbf{ACC} & \textbf{Recall}\\
\hline
LR & Pos & 94.53 & 40.36 & & 89.32 & 61.95 & & 94.54 & 41.53 & & 89.27 & 56.08 \\
LR & Pos+Glove & 94.64 & 52.96 & & 89.36 & 66.84 & & 94.56 & 43.39 & & 89.31 & 60.58 \\
SVM & Pos & 94.55 & 43.19 & & 89.34 & 64.27  & & 94.55 & 42.86 & & 89.28 & 57.14 \\
SVM & Pos+Glove & 94.64 & 53.73 & & 89.36 & 66.58 & & 94.56 & 43.65 & & 89.31 & 60.58 \\
Adaboost & Pos & 94.61 & 50.13 & & 89.35 & 65.04 & & 94.56 & 43.12 & & 89.30 & 58.99\\
Adaboost & Pos+Glove & 94.69 & 58.87 & & 89.42 & 73.78 & & 94.57 & 45.24 & & 89.31 & 60.05 \\
XGBoost & Pos & 94.61 & 49.61 & & 89.38 & 69.15 & & 94.59 & 47.62 & & 89.32 & 62.17 \\
XGBoost & Pos+FreqPhrases & 94.72 & 63.24 & & 89.43 & 74.55 & & 94.61 & 49.47 & & 89.35 & 65.08  \\
XGBoost & Pos+Glove & \textbf{94.74} & \textbf{65.30} & & \textbf{89.46} & \textbf{78.15} & & \textbf{94.63} & \textbf{52.65} & & \textbf{89.36} & \textbf{67.20} \\
\bottomrule
\end{tabular}
\caption{Evaluation results of salient tuple selection. "Pos" denotes positional features, "Glove" is row and column names' Glove embedding, and "FreqPhrases" is the one-hot representation of the fifty most frequent phrases in salient tuples' row and column names. "ACC" and "Recall" are the accuracy and recall of the selected top-n tuples.}
\label{table:tuple_selection}
\end{table*}

As for those thousands of tuples extracted from tables, we regard the salient tuple selection as a binary classification problem. We train and evaluate different classification methods, including the logistic regression (LR), support vector machine (SVM), Adaboost, and XGBoost, on our annotated tuple selection dataset. 
Salient tuples (positive samples) are usually sparse in these report documents. To deal with the class imbalance problem, we perform undersampling over negative samples to ensure the ratio of positive and negative samples is 1:10 in the training set.
Table \ref{table:tuple_selection} shows introducing the word embeddings can benefit recall.
The XGBoost equipped with positional features and Glove embedding \cite{pennington2014glove} outperforms other combinations of classifiers and features, so we use it for the salient tuple selection.

\begin{table}[t]
\small
\renewcommand\arraystretch{1.0}
\centering
\setlength{\tabcolsep}{1.2mm}{
\begin{tabular}{lccccc}
\toprule 
\textbf{Model}& \textbf{Param} & \textbf{\makecell{Enc/Dec\\Layers}} & \textbf{\makecell{Input\\Len}} & \textbf{\makecell{Batch\\Size}} \\
\midrule
\multicolumn{4}{l}{\textbf{Summarizer}}\\
\midrule
$\mathrm{BART}_{large}$ & 406M & 12 & 1,024 & 16 \\
$\mathrm{PEGASUS}_{large}$ & 568M & 16 & 1,024 & 16 \\
$\mathrm{LED}_{large}$ & 460M & 12 & 3,072 & 16 \\
\makecell[l]{BigBird-\\
PEGASUS} & 577M & 16 & 3,072 & 16 \\
\midrule
\multicolumn{4}{l}{\textbf{Tuple-to-Text Generator}}\\
\midrule
$\mathrm{BART}_{base}$ & 139M & 6 & 512 & 8 \\
$\mathrm{BART}_{large}$ & 406M & 12 & 512 & 8 \\
$\mathrm{T5}_{base}$ & 223M & 12 & 512 & 8 \\
$\mathrm{T5}_{large}$ & 737M & 24 & 512 & 8 \\
\bottomrule
\end{tabular}}
\caption{Details of summarizers and text generators.}
\label{table:model_details}
\end{table}

\subsection{Model Details}
\label{appendixsec:model_details}
Table \ref{table:model_details} presents the number of parameters and some hyperparameters of summarization models and tuple-to-text generators used in this work.

\subsection{Evaluation Results}
\label{appendixsec:eval_results}

To handle autoregressive abstractive summarization methods' difficulty in generating long text, we follow a divide-and-conquer method \cite{gidiotis2020divide} and decompose the long summary generation problem into multiple sub-problems of summary segment generation. These summary segments can be generated in parallel and combined as the final summary. 
Table \ref{autoeval:summary_segment} presents ROUGE scores of each output summary segment produced by different models.

\begin{table*}[t]
\renewcommand\arraystretch{1.0}
\small
\centering
\setlength{\tabcolsep}{1.9mm}{
\begin{tabular}{lcccccccccccc}
\toprule 
\multicolumn{4}{l}{\textbf{FINDSum-ROO}}\\
\hline
\textbf{\multirow{2}*{Type}} & \textbf{\multirow{2}*{Method}} & \multicolumn{3}{c}{\textbf{Segment 1}} & &\multicolumn{3}{c}{\textbf{Segment 2}} & & \multicolumn{3}{c}{\textbf{Combined}}\\
\cline{3-5} \cline{7-9} \cline{11-13}
~ & ~ & \textbf{R-1} & \textbf{R-2} & \textbf{R-L} & &\textbf{R-1} & \textbf{R-2} & \textbf{R-L} & & \textbf{R-1} & \textbf{R-2} & \textbf{R-L}\\
\hline
\multirow{6}*{\makecell{Only\\Text}} & LexRank & 34.64 & 8.88 & 16.42 & & 35.73 & 9.76 & 17.20 & & 34.43 & 7.73 & 14.92\\
& TextRank & 35.15 & 9.06 & 16.65 & & 36.00 & 9.79 & 17.20 & & 35.93 & 7.74 & 15.08\\
& BART & 43.13 & 13.82 & 21.10 & & 40.99 & 11.74 & 18.38 & & 49.00 & 16.88 & 19.14 \\
& PEGASUS & 44.79 & 15.17 & 21.53 & & 44.46 & 14.21 & 19.70 & & 51.92 & 19.31 & 21.47 \\
& LED & 46.11 & 16.17 & 22.49 & & 45.52 & 15.17 & 20.26 & & 53.06 & 20.33 & 22.28 \\
& \makecell{BigBird-\\PEGASUS} & 46.25 & 16.78 & 22.67 & & 45.34 & 15.28 & 20.23 & & 53.08 & 20.85 & 20.94\\
\midrule

\multirow{2}*{\makecell{CG}} & CG-LED & 46.99 & 17.44 & 23.14 & & \textbf{47.42} & \textbf{17.04} & \textbf{21.18} & & 54.24 & 22.08 & 23.10 \\
& CG-BigBird & 47.27 & \textbf{18.02} & \textbf{23.24} & & 46.94 & 16.77 & 21.02 & & \textbf{54.40} & \textbf{22.48} & \textbf{23.21}\\
\midrule
\multirow{2}*{\makecell{GCG}} & GCG-LED &  46.98 & 17.23 & 23.06 & &  47.36 & 16.83 & 21.01 & & 54.32 & 21.92 & 23.03 \\
& GCG-BigBird & \textbf{47.28} & 17.87 & 23.15 & & 46.79 & 16.48 & 20.85 & & 54.12 & 22.11 & 23.02 \\
\midrule
\multicolumn{4}{l}{\textbf{FINDSum-Liquidity}}\\
\hline
\textbf{\multirow{2}*{Type}} & \textbf{\multirow{2}*{Method}} & \multicolumn{3}{c}{\textbf{Segment 1}} & & \multicolumn{3}{c}{\textbf{Segment 2}} & & \multicolumn{3}{c}{\textbf{Segment 3}}\\
\cline{3-5} \cline{7-9} \cline{11-13}
~ & ~ & \textbf{R-1} & \textbf{R-2} & \textbf{R-L} & & \textbf{R-1} & \textbf{R-2} & \textbf{R-L} & & \textbf{R-1} & \textbf{R-2} & \textbf{R-L}\\
\hline
\multirow{6}*{\makecell{Only\\Text}} & LexRank & 32.43 & 6.84 & 15.49 & & 32.01 & 6.87 & 15.61 & & 30.64 & 6.06 & 14.67 \\
& TextRank & 33.43 & 7.22 & 15.75 & & 33.09 & 7.09 & 15.79 & & 31.74 & 6.26 & 14.96 \\
& BART & 42.58 & 12.91 & 19.94 & & 39.74 & 10.93 & 18.23 & & 36.99 & 8.31 & 16.34 \\
& PEGASUS &  43.95 & 14.20 & 20.55 & & 40.59 & 11.05 & 18.06 & & 37.25 & 8.80 & 16.40 \\
& LED & 43.49 & 13.37 & 20.03 & & 40.69 & 11.12 & 18.03 & & 38.70 & 9.61 & 16.84 \\
& \makecell{BigBird-\\PEGASUS} & 44.58 & 14.49 & 20.59 & & 40.97 & 11.46 & 18.25 & & 38.16 & 9.63 & 16.87\\
\midrule
\multirow{2}*{\makecell{CG}} & CG-LED & 44.45 & 14.28 & 20.42 & & 41.96 & 12.59 & \textbf{18.89} & & 39.78 & \textbf{10.63} & 17.51 \\
& CG-BigBird &  45.19 & 15.36 & 21.04 & & 41.68 & 12.29 & 18.60 & & 38.45 & 9.92 & 16.96 \\
\midrule
\multirow{2}*{\makecell{GCG}} & GCG-LED & 44.57 & 14.48 & 20.67 & & 41.84 & 12.14 & 18.61 & & \textbf{39.84} & 10.57 & \textbf{17.52} \\
& GCG-BigBird & \textbf{45.34} & \textbf{15.56} & \textbf{21.07} & & \textbf{42.30} & \textbf{12.64} & 18.84 & & 38.73 & 10.35 & 17.22 \\
\bottomrule
\end{tabular}}
\caption{\label{autoeval:summary_segment} Automatic evaluation results of each output summary segment and final combined summary on test sets of FINDSum-ROO and FINDSum-Liquidity.}
\end{table*}

\subsection{Examples of Generation Results}
\label{appendixsec:output_example}

Table \ref{table:case_of_tuple2text_gen} compares different tuple-to-text generation models' outputs and target outputs in an example. 
Table \ref{table:case_of_10k_liquidity} and \ref{table:case_of_10k_roo} show different summarization models' output summaries and the target summary in an example. We show the output summaries' first segments in these two tables.

\begin{table*}[!htb]
\small
\renewcommand\arraystretch{1.2}
\centering
\begin{tabularx}{\textwidth}{lXXX}
\hline
\multicolumn{2}{l}{\textbf{Example of tuple-to-text generation}} \\
\hline
\makecell[l]{Input\\Tuples} & interest bearing deposits with banks |  | 2 \& 2,038 |  , interest bearing deposits with banks |  | 15.7 \& 15,700 |  , interest bearing deposits with banks \& federal funds sold and securities purchased under agreements to resell |  | 17.8 \& 17,838 |  , interest bearing deposits with banks \& federal funds sold and securities purchased under agreements to resell |  | 10.2 \& 10,168 |  , ficial assets \& short-term ficial assets \& federal funds sold and securities purchased under agreements to resell |  | 17.8 \& 17,838 |  , ficial assets \& short-term ficial assets \& federal funds sold and securities purchased under agreements to resell |  | 10.2 \& 10,168 | total short-term borrowings |  | 3.7 \& 3,659 | 2019\\
\hline
\makecell[l]{Target\\Output} & interest bearing deposits with banks totaled \$ 2,038 million and \$ 15,700 million at december 31. federal funds sold and securities purchased under agreements to resell totaled \$ 17,838 million and \$ 10,168 million at december 31. short-term borrowings totaled \$ 3,659 million at december 31.\\
\hline
\makecell[l]{BART-\\large} & interest bearing deposits with banks totaled \$ 2,038 million at december 31, 2019, compared to \$ 15,700 million at year-end 2018. federal funds sold and securities purchased under agreements to resell totaled \$ 17,838 million and \$ 10,168 million at the end of 2019 and 2018. short-term borrowings totaled \$ 3,659 million.\\
\hline
\makecell[l]{BART-\\base} & interest bearing deposits with banks totaled \$ 2.0 billion at december 31, 2019, an increase of \$ 15.7 billion, or 8.1 percent, at an average rate of 4.8 percent. federal funds sold and securities sold under agreements to resell totaled \$ 17.8 billion. short-term borrowings held at the federal fhl and fhl totaled \$ 3.2 billion. the federal funds held and securities held with banks held with the fhl increased totaled \$ 1.0 million, or 3.7 percent, were an increase in federal and flo borrowings.\\
\hline
T5-large & federal funds sold and securities purchased under agreements to resell were \$ 17.8 million and \$ 10.2 million for the years ended december 31, 2019 and 2018, respectively. short-term borrowings totaled \$ 3.7 million.\\
\hline
T5-base & interest bearing deposits with banks totaled \$ 2,038 million at december 31, 2019, compared to \$ 15,700 million, \$ 17,838 million and \$ 10,168 million in 2019 and 2018. short-term borrowings were \$ 3,659 million in 2018, an increase of \$ 2 million from 2018. the increase was primarily attributable to a \$ 3.7 million increase in borrowings under our revolving credit facility, partially offset by an increase in federal funds sold and securities purchased under agreements to resell of \$ 17.8 million and a decrease in short term borrowings of \$ 15.7 million.\\
\hline
\end{tabularx}
\caption{An example of tuple-to-text generation results.}\label{table:case_of_tuple2text_gen}
\end{table*}

\begin{table*}[!htb]
\small
\renewcommand\arraystretch{1.1}
\centering
\begin{tabularx}{\textwidth}{lXXX}
\hline
\multicolumn{2}{l}{\textbf{Example from the FINDSum-Liquidity}} \\
\hline
\makecell[l]{Target\\Summary} & 
the following table sets forth a summary of the primary sources and uses of cash ( in thousands ) : cash used in operating activities during the year ended december 31 , 2019 , cash used in operating activities was \$ 85.0 million , which consisted of a net loss of \$ 94.4 million , adjusted by non-cash charges of \$ 15.1 million and cash used due to changes in our operating assets and liabilities of \$ 5.7 million . the non-cash charges consisted primarily of depreciation and amortization expense of \$ 3.4 million , stock-based compensation of \$ 5.3 million , and non-cash operating lease expense of \$ 6.4 million . the change in our operating assets and liabilities was primarily due to a decrease of \$ 4.3 million as a result of the deferred revenue recorded in connection with our collaboration agreement with bluebird , a decrease of \$ 3.3 million due to pre-payments made per our lease agreements , and a decrease of \$ 0.3 million in deposits and other long-term assets , offset by increases of \$ 1.5 million in accrued research and development expenses and \$ 0.7 million in accrued compensation . during the year ended december 31 , 2018 , cash used in operating activities was \$ 38.2 million , which consisted of a net loss of \$ 64.8 million , adjusted by non-cash charges of \$ 6.4 million and cash used due to changes in our operating assets and liabilities of \$ 20.2 million . the non-cash charges consisted primarily of depreciation and amortization expense of \$ 3.4 million and stock-based compensation of \$ 3.0 million .\\
\hline
Model Output&\\
\hline
GCG-Bigbird & the following table sets forth the primary sources and uses of cash for each of the periods set forth below ( in thousands ) : operating activities net cash used in operating activities was \$ 85.0 million for the year ended december 31, 2019, primarily resulting from our net loss of \$ 94.4 million, which was partially offset by non-cash charges of \$ 50.0 million for depreciation and amortization, \$ 0.7 million for stock-based compensation, and \$ 6.4 million for operating lease expense. the changes in our operating assets and liabilities were primarily due to an increase in prepaid expenses and other current assets of \$ 2.5 million and a decrease in accounts payable of \$ 41.8 million and accrued research and development expenses of \$ 1.5 million. these increases were offset by decreases in deferred revenue of \$ 5.5 million and accounts receivable of \$ 1.2 million and an increase of \$ 0.4 million in other non-current assets. net cash ( used in ) provided by operating activities of \$ 38.2 million was primarily the result of \$ 75.2 million in net proceeds from our ipo, net of underwriting discounts and commissions and offering costs, as well as cash proceeds of \$ 69.7 million from our at-the-market offering and \$ 60.0 million from the sale of our common stock under the sales agreement with cowen, offset in part by a \$ 15.8 million increase in cash used for investing activities, including purchases of property and equipment of \$ 5.7 million and purchases of available-for-sale marketable securities of \$ 80.0 million and cash outflows of \$ 12.0 million related to changes in the fair value of our marketable securities.\\
\hline
CG-Bigbird & since our inception, we have incurred significant losses and negative cash flows from operations. we have an accumulated deficit of \$ 221.0 million through december 31, 2019. we expect to incur substantial additional losses in the future as we conduct and expand our research and development activities and invest in our manufacturing facility. cash flows the following table sets forth the primary sources and uses of cash for each of the periods set forth below ( in thousands ) : operating activities net cash used in operating activities was \$ 81.0 million for the year ended december31, 2019, and consisted primarily of a net loss of \$ 94.4 million, partially offset by non-cash charges of \$ 55.0 million and net cash provided by changes in our operating assets and liabilities of \$ 15.8 million. the net loss was primarily due to an increase in the net deferred revenue of \$ 4.3 million due to the timing of payments, a decrease in accounts payable of \$ 41.8 million as a result of timing of cash receipts, an increase of \$ 1.5 million in accrued expenses and other current assets, offset by a decrease of \$ 1.6 million in accounts receivable due to timing of collections, \$ 5.3 million in stock-based compensation expense and \$ 6.4 million in operating lease expense related to the operating lease of our new corporate headquarters building. these decreases in cash used were partially offset, in part, by a net increase in deferred revenues of \$ 7.3 million due primarily to the recognition of milestone payments under our collaboration agreement with bluebird bio, inc., or bluebird, of \$ 2.5 million and a net decrease in accrued compensation expense of \$ 0.5 million related to our former chief executive officer.
\\
\hline
\makecell[l]{BigBird-\\PEGASUS} & since our inception, we have incurred significant losses and negative cash flows from operations. we have funded our operations to date primarily from private placements of our convertible preferred stock, the net proceeds from our initial public offering, or ipo, which we completed in october 2018, from our follow-on public offering ( which was completed in april 2019 ), as well as cash proceeds from bluebird under the collaboration agreement we entered into in august 2018. we expect to continue to incur net operating losses for at least the next several years as we advance our personalized cancer immunotherapy through clinical development, seek regulatory approval, prepare for and, if approved, proceed to commercialization, continue our research and development efforts and invest in our manufacturing facility. cash flows the following table sets forth the primary sources and uses of cash for each of the periods set forth below : operating activities during the year ended december 31, 2019, net cash used in operating activities was \$ 64.6 million, primarily resulting from our net loss of \$ 92.2 million and changes in our operating assets and liabilities, partially offset by non-cash charges totaling \$ 19.9 million. the net loss was primarily due to the costs incurred in connection with the development of our slate, granite and bisab product candidates and general and administrative expenses associated with our operations, including the costs associated with being a public company.
\\
\hline
\end{tabularx}
\caption{An example of output summaries from the FINDSum-Liquidity.}\label{table:case_of_10k_liquidity}
\end{table*}

\begin{table*}[!htb]
\small
\renewcommand\arraystretch{1.1}
\centering
\begin{tabularx}{\textwidth}{lXXX}
\hline
\multicolumn{2}{l}{\textbf{Example from the FINDSum-ROO}} \\
\hline
\makecell[l]{Target\\Summary} & the following table sets forth the percentage relationships of expense items to total operating revenue for the periods indicated: year ended december 31 , 2012 compared with the year ended december 31 , 2011 operating revenue increased \$ 17.1 million ( 3.2 \% ) , to \$ 545.7 million for the year ended december 31 , 2012 from \$ 528.6 million for the year ended december 31 , 2011. the increase in revenue was the result of a \$ 4.6 million ( 4.3 \% ) increase in fuel surcharge revenue from \$ 107.8 million in 2011 to \$ 112.4 million in 2012 and an increase in line haul and other revenues of \$ 12.5 million ( 3.0 \% ). fuel surcharge revenues represent fuel costs passed on to customers based on customer specific fuel charge recovery rates and billed loaded miles. fuel surcharge revenues increased mostly as a result of a 3.2 \% increase in average doe diesel fuel prices during the year ended december 31 , 2012 compared to the same period of 2011. line haul revenues increased mainly as a result of freight rate increases offset by a slight decrease in loaded miles. salaries , wages , and benefits increased \$ 0.4 million ( 0.2 \% ) , to \$ 167.1 million for the year ended december 31 , 2012 from \$ 166.7 million in the 2011 period. the increase was the result of a \$ 0.8 million increase ( 0.7 \% ) in driver wages , a \$ 2.2 million increase in amortization of stock-based compensation awards , a \$ 2.7 million ( 38.5 \% ) decrease in workers' compensation , and a \$ 0.1 increase in other compensation and benefits. the company driver wage increase was mainly due an increase in miles driven.\\
\hline
Model Output & \\
\hline
GCG-Bigbird & years ended december 31, 2012, 2011, and 2010 operating revenue operating revenue increased \$ 17.1 million ( 17.1 \% ), to \$ 545.7 million for 2012 from \$ 528.0 million for 2011. the increase in revenue was mainly the result of an increase in fuel surcharge revenue from \$ 107.8 million in 2011 to \$ 102.5 million in 2012. line haul and other revenues decreased \$ 3.4 million ( 0.8 \% ) on a decrease in total miles, offset by a increase in freight rates per total mile. fuel surcharge revenues increased mostly as a result of a 28.4 \% increase in average doe diesel fuel prices during the year ended december 2011 compared to the same period of 2010, as well as an improvement in loaded miles versus total miles comparing the two periods. salaries, wages and benefits salaries and benefits increased \$ 1.3 million ( 0.6 \%, ) to \$ 167.1 million for the year 2012 from the year 2011 primarily due to an increase of \$ 2.1 million in driver wages, a \$ 0.4 million increase in office and shop wages and a \$ 1.0 million increase of workers'compensation. rent and purchased transportation rent and purchase transportation decreased \$ 1.9 million ( 20.4 \% ), to \$ 3.0 million for 2011 from \$ 9.5 million in the 2010 period. the decrease is mainly attributable to fewer independent contractors driving for the company. during the 2011 period, independent contractors accounted for 2 \% of the total fleet miles compared to approximately 3 \% for the comparable period of 2010. the company's driver wage decrease was mainly due to fewer miles driven year over year, which was partially offset by an overall decrease in the overall company average wage rate per mile due mainly to the mix of company drivers based on operating region locations.
\\
\hline
CG-Bigbird & the following table sets forth, for the periods indicated, the percentage that certain items in the statements of operations bear to operating revenue. year ended december 31, 2012 compared with the year ended january 1, 2011 operating revenue increased \$ 30.0 million ( 17.0 \% ), to \$ 540.0 million for the yearended december 31 2012 from \$ 528.6 million for 2011. the increase in revenue was mainly the result of a \$ 32.5 million ( 43.2 \% fuel surcharge ) increase in fuel surcharge revenue from \$ 107.8 million in 2011 to \$ 107.7 million in 2012. line haul and other revenues decreased \$ 0.4 million ( 0.8 \% ) on a decrease in total miles, offset by an increase in freight rates per total mile. fuel surcharge revenues represent fuel costs passed on to customers based on customer specific fuel charge recovery rates and billed loaded miles. there was an improvement in loaded miles versus total miles compared to the same period in 2011. fuel cost per mile, net of fuel surcharge, increased 14.9 \% in the 2012 period compared to 2011, as a result of increased fuel prices, \$ 34.4 million, which was offset by a decrease of \$ 20.6 million in volume. salaries, wages, and benefits increased \$ 0.9 million ( 0.9 \%, primarily due to a \$ 0.6 million increase in non-driver personnel wages. property and land improvements decreased \$ 0.1 million ( 0.1 \%. ) insurance and claims decreased \$ 0.2 million ( 0.2 \% ; ) rent and purchased transportation decreased \$ 0.3 million ( 0.3 \% ). the decrease is mainly attributable to lower amounts paid to independent contractors due to fewer miles driven. during the fourth quarter of 2012, independent contractors accounted for 2 \% of the total fleet miles compared to approximately 3 \% for the same quarter in 2011. \\
\hline
\makecell[l]{BigBird-\\PEGASUS} & the following table sets forth, for the periods indicated, the percentage of total revenues represented by certain items reflected in the company's consolidated statements of income. year ended december 31, 2012 compared with the year ended january 1, 2011 operating revenue increased \$ 17.8 million ( 14.9 \% ), to \$ 528.6 million for the yearended december 31, 2012 from \$ 528.8 million for 2011. the increase in revenue was mainly the result of a \$ 32.5 million ( 43.2 \%, fuel surcharge revenues ) increase in fuel surcharge revenue from \$ 75.3 million in 2011 to \$ 107.8 million in 2012, offset by a decrease in line haul and other revenues of \$ 3.4 million, or 0.8 \%. operating expenses, net, increased \$ 8.7 million ( 9.9 \% , excluding the impact of the fuel surcharge increase, which was \$ 11.9 million ) to \$ 284.8 million for 2012 from operating expenses of \$ 264.5 million for 2011. the increase is mainly attributable to a \$ 18.6 million ( 21.4 \% ), or \$ 13.0 million increase in salaries, wages, and benefits, primarily due to an increase in non-driver personnel in 2012 compared to the same period in 2011, as well as a \$ 2.7 million increase in other benefits and payroll taxes, mainly due to a higher percentage of other benefits paid to independent contractors as a result of fewer independent contractors driving for the company, partially offset by an increase of \$ 2.2 million in rent and purchased transportation, a decrease of \$ 1.9 million in workers'compensation and a \$ 1.0 million decrease in health insurance, both of which were due to frequency and severity of claims. depreciation decreased \$ 4.7 million ( 7.6 \% decrease in depreciation expense, which is primarily attributable to the decrease in average depreciation per tractor.
\\
\hline
\end{tabularx}
\caption{An example of output summaries from the FINDSum-ROO.}\label{table:case_of_10k_roo}
\end{table*}

\end{document}